\documentclass{article}

\PassOptionsToPackage{numbers, sort&compress}{natbib}
\usepackage{subcaption,booktabs}
\usepackage{graphicx}

\usepackage{xspace}
\usepackage{tikz}
\usetikzlibrary{fit, positioning, shapes,arrows}
\usepackage{amsmath}
\usepackage{amssymb}
\usepackage{float}
\usepackage{multirow}
\newcommand\numberthis{\addtocounter{equation}{1}\tag{\theequation}}
\usepackage[tableposition=top]{caption}
\usepackage{multicol}
\DeclareCaptionLabelFormat{andtable}{#1~#2  \&  \tablename~\thetable}
\usepackage{xfrac}
\usepackage{wrapfig}
\usepackage{cutwin,lipsum}
\usepackage{picinpar}
\tikzset{three sided/.style={
		draw=none,
		append after command={
			[shorten <= -0.5\pgflinewidth]
			([shift={(-1.5\pgflinewidth,-0.5\pgflinewidth)}]\tikzlastnode.north east)
			edge([shift={( 0.5\pgflinewidth,-0.5\pgflinewidth)}]\tikzlastnode.north west) 
			([shift={( 0.5\pgflinewidth,-0.5\pgflinewidth)}]\tikzlastnode.north west)
			edge([shift={( 0.5\pgflinewidth,+0.5\pgflinewidth)}]\tikzlastnode.south west)            
			([shift={( 0.5\pgflinewidth,+0.5\pgflinewidth)}]\tikzlastnode.south west)
			edge([shift={(-1.0\pgflinewidth,+0.5\pgflinewidth)}]\tikzlastnode.south east)
		}
	}
}

\makeatletter
\newcommand*\bigcdot{\mathpalette\bigcdot@{.5}}
\newcommand*\bigcdot@[2]{\mathbin{\vcenter{\hbox{\scalebox{#2}{$\m@th#1\bullet$}}}}}
\makeatother

\usepackage[export]{adjustbox}
\makeatletter
\def\algbackskip{\hskip-\ALG@thistlm}
\makeatother

\usepackage{dsfont}

\usepackage{wasysym}
\usetikzlibrary{calc,positioning}

\usepackage{booktabs}
\usepackage{multirow}

\usepackage{enumitem}

\PassOptionsToPackage{numbers, sort&compress}{natbib}
\usepackage[preprint]{neurips_2019}
\usepackage[utf8]{inputenc} % allow utf-8 input
\usepackage[T1]{fontenc}    % use 8-bit T1 fonts
\usepackage{hyperref}       % hyperlinks
\usepackage{url}            % simple URL typesetting
\usepackage{booktabs}       % professional-quality tables
\usepackage{amsfonts}       % blackboard math symbols
\usepackage{nicefrac}       % compact symbols for 1/2, etc.
\usepackage{microtype}      % microtypography

\author{%
	 Virginia Aglietti \\
	University of Warwick\\
	The Alan Turing Institute\\
	\texttt{V.Aglietti@warwick.ac.uk} \\
	\And
	Edwin V. Bonilla \\
	CSIRO's Data61\\
	UNSW \\
	\texttt{Edwin.Bonilla@data61.csiro.au} \\
	\And
	Theodoros Damoulas \\
	University of Warwick\\
	The Alan Turing Institute\\
	\texttt{T.Damoulas@warwick.ac.uk} \\		
	\And 
	Sally Cripps\\
	Centre for Translational Data Science\\
	The University of Sydney\\
	\texttt{Sally.Cripps@sydney.edu.au} \\		
}

\newcommand{\ie}{i.e.\xspace}
\newcommand{\eg}{e.g.\xspace}

\newcommand{\eq}{Eq.\xspace}
\newcommand{\eqs}{Eqs.\xspace}
\newcommand{\fig}{Fig.\xspace}

\newcommand{\tbl}{Tab.\xspace}

\newcommand{\acro}[1]{\textsc{#1}\xspace}
\newcommand{\lgcp}{\acro{lgcp}}

\newcommand{\cmcp}{\acro{cmcp}}

\newcommand{\sgcp}{\acro{sgcp}}
\newcommand{\gptext}{\acro{gp}}

\newcommand{\mcmc}{\acro{mcmc}}

\newcommand{\nlpl}{\acro{nlpl}}

\newcommand{\cpu}{\acro{cpu}}

\newcommand{\termelbo}{\acro{elbo}}

\newcommand{\lemma}{\acro{lemma}}

\newcommand{\adams}{\acro{sgcp}}
\newcommand{\donner}{\acro{mfvb}}
\newcommand{\lloyds}{\acro{vbpp}}
\newcommand{\our}{\acro{stvb}}

\newcommand{\vi}{\acro{vi}}
\newcommand{\mf}{\acro{mf}}

\newcommand{\ppp}{\acro{ppp}}
\newcommand{\svi}{\acro{svi}}
\newcommand{\coverage}{\acro{ec}}
\newcommand{\credible}{\acro{ci}}

\newcommand{\defeq}{\stackrel{\text{\tiny def}}{=}}
\newcommand{\mat}[1]{\mathbf{#1}}
\renewcommand{\vec}[1]{ \mathbf{#1} } % math bold
\newcommand{\vecS}[1]{\boldsymbol{ #1 }  } % this for boldsymbols

\newcommand{\gp}{\mathcal{GP}}
\newcommand{\kernel}{\kappa}

\newcommand{\covfunc}[3]{\kernel(#1,#2; #3)}
\newcommand{\hyperparam}{\vectheta}

\newcommand{\x}{\vec{x}}
\newcommand{\y}{\vec{y}}

\newcommand{\xprime}{\x^{\prime}}
\newcommand{\vectheta}{\vecS{\theta}}

\newcommand{\g}{\mat{G}}

\newcommand{\f}{\vec{f}}

\renewcommand{\u}{\vec{u}}

\newcommand{\z}{\vec{z}}
\newcommand{\Z}{\mat{Z}}

\newcommand{\Kv}{\mat{K}_{xx}}

\newcommand{\Kzz}{\mat{K}_{zz}}
\newcommand{\Kxz}{\mat{K}_{xz}}
\newcommand{\Kzx}{\mat{K}_{zx}}
\newcommand{\Kzzinv}{(\mat{K}_{zz})^{-1}}

\newcommand{\A}{\mat{A}}

\newcommand{\normal}{\mathcal{N}}

\newcommand{\expectation}[2]{ \mathbb{E}_{#1}{\left[#2\right]} }

\newcommand{\varmeanu}{\vec{m}}

\newcommand{\varcovu}{\mat{S}}

\newcommand{\calL}{\mathcal{L}}
\newcommand{\elbo}{\calL_{\text{elbo}}}

\newcommand{\klterm}{\calL_{\text{kl}}}
\newcommand{\enterm}{\calL_{\text{ent}}}

\newcommand{\lambdamax}{\lambda^\star}
\newcommand{\region}{\mathcal{X}}

\newcommand{\parameterM}{\mu(\u)}

\newcommand{\lnorm}{\mathit{l}_2}
\newcommand{\ltest}{\ell_{test}}

\title{Structured Variational Inference in \\ Continuous Cox Process Models}

\begin{document}

\maketitle

\begin{abstract}
We propose a scalable framework for inference in an inhomogeneous Poisson process modeled by a continuous sigmoidal Cox process that assumes the corresponding intensity function is given by a Gaussian process (\gptext) prior  transformed with a scaled logistic sigmoid function.  We present a tractable representation of the likelihood through augmentation with a  superposition of Poisson processes. This view enables a structured variational approximation capturing dependencies across variables in the model. Our framework avoids discretization of the domain, does not require accurate numerical integration over the input space and is not limited to \gptext{s} with squared exponential kernels. We evaluate our approach on synthetic and real-world data showing that its benefits are particularly pronounced on multivariate input settings where it overcomes the limitations of  mean-field methods and sampling schemes. We provide the state of-the-art in terms of speed, accuracy and uncertainty quantification trade-offs.
\end{abstract}

\section{Introduction}

Point processes have been used effectively to model a variety of event data such as occurrences of diseases \cite{diggle2013spatial, lloyd-icml-2015}, location of earthquakes \cite{marsan2008extending} or crime events \cite{grubesic2008spatio, aglietti2018} . The most commonly adopted class of models for such discrete data are non-homogenous Poisson processes and in particular Cox processes \cite{cox1955some}. In these, the observed events are assumed to be generated from a Poisson point process (\ppp) whose intensity is stochastic, enabling non-parametric inference and uncertainty quantification. 

Gaussian processes \citep[\gptext{s};][]{Rasmussen:2005:GPM:1162254} form a flexible prior over functions and, therefore, have been used to model the intensity of a Cox process via a non-linear positive link function. Typical mappings are the exponential \cite{moller1998log, diggle2013spatial}, the square \cite{lloyd-icml-2015,pmlr-v37-lian15} and the sigmoidal \cite{adams2009tractable, gunter2014efficient, donner2018efficient} transformations.  In general, inferring the intensity function over a continuous input space $\region$ is highly problematic and different algorithms have been proposed to deal with this issue depending on the transformation used. For example, under the exponential transformation, a regular computational grid is commonly introduced \cite{diggle2013spatial}. While this significantly simplifies inference, it leads to poor approximations, especially in high dimensional settings. Increasing the resolution of the grid to improve the approximation  yields computationally prohibitive algorithms that do not scale, highlighting the well-known trade-off between statistical performance and computational cost. 

Other algorithms have been proposed to deal with a continuous $\region$ but they are computationally expensive \cite{adams2009tractable, gunter2014efficient}, are limited to simple covariance functions \cite{lloyd-icml-2015}, require accurate numerical integration over the domain \cite{donner2018efficient} or do not account for the model dependencies in the posterior distribution \cite{donner2018efficient}.  In this paper we propose an inference framework that addresses all of these modeling and inference limitations by having a tractable representation of the likelihood via augmentation with a superposition of \ppp{s}. This enables a scalable structured variational inference algorithm (\svi) in the continuous space directly, where the approximate posterior distribution incorporates dependencies between the variables of interest. Our specific contributions are as follows.

\textbf{Scalable inference in continuous input spaces}: The augmentation of the input space via a process superposition view allows us to develop a scalable variational inference algorithm that does not require discretization or \emph{accurate} numerical integration. With this view, we obtain a joint distribution that is readily normalized, providing a natural regularization over the latent variables in our model.

\textbf{Efficient structured posterior estimation}: We estimate a joint posterior  that captures the complex variable dependencies in the model while being significantly faster than sampling approaches. 
	
\textbf{State-of-the-art performance}: Our experimental evaluation shows the benefits of our approach when compared to state-of-the-art  inference schemes, link functions, augmentation schemes and representations of the input space $\mathcal{X}$.

% Approximated (discretisation-based) methods have as its
%main disadvantages: i) producing biased results; ii) it is hard (sometimes impossible) to
%quantify the bias (error) involved; iii) the level of discretisation required to obtain good
%(close enough to the exact) results is unknown and case-specific; iv) it is not always clear
%to which limit these approaches converge to. This motivates the development of exact
%methodologies, i.e. free from discretisation errors and helps to understand its advantages.
%no discretisation-based approximation is used

\subsection{Related work}\label{sec:related_work}
% LGCPS and exponential
\gptext-modulated point processes are the gold standard for modeling event data. Performing inference in these models, \eg under the exponential transformation, has typically required discretization where the domain is gridded and the intensity function is assumed to be constant over each grid cell \cite{moller1998log, cunningham2008fast, diggle2013spatial,brix2001spatiotemporal}. Alternatively, \citet{lasko2014efficient} also considers an exponential link function and performs inference over a renewal process resorting to numerical integration  within a computationally expensive sampling scheme. 
%which scales as $\mathcal{O}(k^3) + \mathcal{O}(N)$ with $k$ denoting the number of integration points.   
These methods  suffer from poor scaling with the dimensionality of $\region$ and sensitivity to the choice of discretization or numerical integration technique. Several approaches have been proposed to deal with inference in the continuous domain directly by using alternative transformations along with additional modeling assumptions and computational tricks or by constraining the \gptext \cite{lopez2019gaussian}. 

% PERMANENTAL (squared)
One of those alternative transformation is the squared mapping as developed in the Permanental process \citep{pmlr-v70-walder17a,lloyd-icml-2015,pmlr-v37-lian15, lloyd2016latent, john2018large}. Although the square transformation enables analytical computation of the required integrals over $\region$, this only holds for certain standard types of kernels such as the squared exponential. In addition, Permanental processes suffer from important identifiability issues such as reflection invariance and lead to model with ``nodal lines'' \cite{john2018large}.  

% Sigmoidal (adams)
Another transformation is the scaled logistic sigmoid function proposed by \cite{adams2009tractable}, which exploits augmentation of the input space via thinning \cite{lewis1979simulation} to achieve tractability. This model is known as the sigmoidal Gaussian Cox process (\sgcp). Their proposed inference algorithm is based on Markov chain Monte Carlo (\mcmc), which enables drawing `exact' samples from the posterior intensity. However, as acknowledged by the authors,  it has significant computational demands making it inapplicable to large datasets. As an extension to this work,  \cite{gunter2014efficient} introduce the concept of ``adaptive thinning'' and propose an expensive \mcmc scheme which scales as $\mathcal{O}(N^3)$.
% better sigmoidal (donner)
More recently, \cite{donner2018efficient} introduced a neat double augmentation scheme for \sgcp which enables closed form updates using a mean-field approximation (\vi-\mf).  However, it requires accurate numerical integration over $\region$, which makes the performance of the algorithm highly dependent on the number of integration points. 

% why we are better
In this work, we overcome the limitations of the mentioned \vi-\mf and \mcmc schemes by proposing an \svi framework, henceforth \our, which takes into account the complex posterior dependencies while being scalable and thus applicable to high-dimensional real-world settings. 
%Our work is thus also related to the literature on \svi \cite{hoffman2015structured} in the context of non-Gaussian likelihood functions \cite{linderman2015scalable, ko2012large}. Variational inference for Cox processes has so far only used \vi-\mf and favored speed over accuracy and uncertainty quantification. Our proposed algorithm balances these three aspects providing higher accuracy while correctly quantifying uncertainty and being faster than \mcmc schemes. 
To the best of our knowledge we are the first to propose a fast structured variational inference framework for \gptext modulated point process models.  
See \tbl \ref{tab:summary_related} for a summary of the most relevant related works.

\begin{table}
	\caption{Summary of related work. $\int$ and $\sum$ denote continuous and discrete models respectively. $M$ represents the number of thinned points derived from the thinning of a \ppp. $K$ indicates the number of inducing inputs. }
	\label{tab:summary_related}
	\begin{center}
		\resizebox{\columnwidth}{!}{\begin{tabular}{lccccccccc}
				  & \textbf{\our} & \lgcp \cite{moller1998log}  & \adams \cite{adams2009tractable} & \citet{gunter2014efficient} &\lloyds \cite{lloyd-icml-2015}& \citet{pmlr-v37-lian15} &  \donner \cite{donner2018efficient}  \\
				\toprule
				\textbf{Inference}    & \svi & \mcmc & \mcmc & \mcmc  & \vi-\mf & \vi-\mf & \vi-\mf    \\
				\textbf{$\mathcal{O}$}    &  $K^3$ & $N^3$ & $(N+M)^3$ & $(N+M)^3$  & $NK^2$ & $NK^2$ & $NK^2$\\
				\pmb{$\lambda(x)$}  &  $\lambdamax\sigma(f(x))$ &   $\exp(f(x))$ &    $\lambdamax\sigma(f(x))$  & $\lambdamax\sigma(f(x))$   &   $(f(x))^2$ & $(f(x))^2$ &   $\lambdamax\sigma(f(x))$  \\
				\pmb{$\region$}  & $\int$ &$\sum$ & $\int$ & $\int$  &  $\int$ & $\sum$ &$\int$\\
				\bottomrule
		\end{tabular}}  
	\end{center}
\end{table}

%Variational inference for the CGP setting has so far only used the mean-field approximation as
%described in [5]. When posterior dependencies are a quantity of interest, a natural approach is to
%increase the complexity of the variational posterior to capture these dependencies. This often results
%in a prohibitive increase in the complexity of the inference.

\section{Model formulation}
We consider learning problems where we are given a dataset of $N$ events  $\mathcal{D} = \{\x_n\}_{n=1}^{N}$, where $\x_n$  is a $d$-dimensional vector in the compact space $\region \subset \mathbb{R}^D$. We aim at modeling these data via a \ppp, inferring the latent intensity function $\lambda(\x):\region \to \mathbb{R}^+$ and making probabilistic predictions. 

\subsection{Sigmoidal Gaussian Cox process} 
Consider a realization $\xi = (N, \{\x_1,...,\x_n\})$ of a \ppp on $\region$ where the points $\{\x_1, ..., \x_n\}$ are treated as \textit{indistinguishable} apart from their locations \cite{daley2007introduction}. 
Conditioned on $\lambda(\x)$, the Cox process likelihood function evaluated at $\xi$ can be written as: 
\begin{align}
	\mathcal{L}(\xi|\lambda(\x)) = \exp\left(-\int_{\tau}\lambda(\x)d\x\right) \prod_{n=1}^{N}\lambda(\x_n) ,
\label{eq:intractable_likelihood}
\end{align}
where the intensity is given by $\lambda(\x) = \lambdamax\sigma(f(\x))$ with $\lambdamax >0$ being an upperbound on $\lambda(\x)$,  $\sigma(\cdot)$ denoting the the logistic sigmoid function and $f$ is drawn from a zero-mean \gptext prior with covariance function $\covfunc{\x}{\xprime}{\hyperparam}$ and hyperparameters $\hyperparam$, \ie $f | \hyperparam \sim \gp(\vec{0}, \covfunc{\x}{\xprime}{\hyperparam})$.  We will refer to this joint model as the sigmoidal Gaussian Cox process (\sgcp ).  Notice that, when considering the tuple $(\x_1, ..., \x_n)$ instead of the set $\{\x_1, ..., \x_n\}$, and thus the event $\xi_0= (N, (\x_1,...,\x_n))$, the likelihood function is given by  $\mathcal{L}(\xi_0|\lambda(\x))  = \frac{\mathcal{L}(\xi|\lambda(\x))}{N!}$. There are indeed $N!$ permutations of the events $\{\x_1,...,\x_n\}$ giving the same point process realization. When the set $\{\x_1, ..., \x_n\}$ is known, considering $\mathcal{L}(\xi|\lambda(\x))$ or $\mathcal{L}(\xi_0|\lambda(\x))$ does not affect the inference procedure. The same holds for \mcmc algorithms inferring the event locations. In this case, the factorial term disappears in the computation of the acceptance ratio. However, as we shall see later, when the event locations are latent variables in a model and inference proceeds via a variational approximation the difference between the two likelihoods is essential. Indeed, while $\mathcal{L}(\xi_0|\lambda(\x))$ is normalized with respect to $N$, one must be cautious when integrating the likelihood in \eq \ref{eq:intractable_likelihood} over sets and bring back the missing $N!$ factor so as  to obtain a proper discrete probability mass function for $N$. 

As it turns out, inference in \sgcp is \textit{doubly intractable}, as it requires solving the  integral in \eq \eqref{eq:intractable_likelihood} and then computing the intractable  posterior distribution $p(\f_N, \lambdamax| \{\x_n\}_{n=1}^{N})$, where $\f_N$ denotes the latent function at the $N$ event locations. One way to achieve tractability is through augmentation of the input space \cite{adams2009tractable,donner2018efficient}, a procedure that introduces precisely those latent (event) variables that require explicit normalization during variational inference.  We will describe below a process superposition view of this augmented scheme that allows us to define a proper distribution over the joint space of observed and latent variables and carry out posterior estimation via variational inference.

%We thus augment \eq \ref{eq:intractable_likelihood} via the superposition of \ppp which leads to a joint distribution over tuples and in turns ensure proper normalisation. 

\subsection{Augmentation via superposition}
A very useful property of independent \ppp{s} is that their superposition, which is defined as the combination of events from two processes in a single one, is a \ppp. Consider two \ppp with intensities $\lambda(\x)$ and $\nu(\x)$ and realisations $(N, \{\x_1, ...,\x_n \})$ and $(M, \{\y_1, ..., \y_M\})$ 
respectively. The combined event $\xi_R = (R = M+N, \{\vec{v}_1, ..., \vec{v}_{R}\})$ is a realization of a \ppp with intensity given by $\lambda(\x) + \nu(\x))$ where knowledge of which points originated from which process is assumed lost. The likelihood for $\mathcal{L}(\xi_R|\lambda(\x),\nu(\x))$ can be thus written as:
\begin{equation}
	\sum_{N=0}^{R} \binom{N + M}{N} \sum_{P_N \in \mathbb{P}_N}\left(\frac{\exp(-\int_{\region} \lambda(\x)d\x)}{N!}\prod_{r \in P_N}\lambda(r)\times\frac{\exp(-\int_{\region} \nu(\x)d\x)}{M!}\prod_{r \in P_N^c}\nu(r)\right),
\end{equation}
where $\mathbb{P}_N$ denotes the collection of all possible partitions of size $N$, $P_N$ represents an element of $\mathbb{P}_N$ and $P_N^c$ is its complement.

Consider now $R=N+M$ to be the total number of events resulting from thinning \cite{lewis1979simulation} where $N$ is the number of observed events while $M$ is the number of latent events with stochastic locations $\y_1, ..., \y_M$. We assume that the probability of observing an event is given by $\sigma(f(\x))$ while the probability for the event to be latent is $\sigma(-f(\x))$. In addition, let $\lambdamax\int_{\region}d\x$ be the expected total number of events. We can see the realization $(M+N, (\x_1, ..., \x_N, \y_1, ..., \y_M))$ as the result of the superposition of two \ppp{s} with intensities $\lambda(x) = \lambdamax\sigma(f(\x))$ and $\nu(\x) = \lambdamax\sigma(-f(\x))$. Differently from the standard superposition, we do know which events are observed and which are latent. In writing the likelihood for $(M+N, \{\x_1, ..., \x_N, \y_1, ..., \y_M\})$ we thus do not need to consider all the possible partitions of $N$. We can write $\mathcal{L}_{N+M}   \defeq  \mathcal{L}(N+M, (x_1, ..., x_N, y_1, ..., y_M))$:
\begin{align}
\mathcal{L}_{N+M}  
	&= \frac{\exp(-\int_{\region} \lambda(\x)d\x)}{N!}\prod_{r \in P_N}\lambda(r)\times\frac{\exp(-\int_{\region} \nu(\x)d\x)}{N!}\prod_{r \in P_N^c}\nu(r) \\
	&= \frac{1}{N!M!}\exp(-\lambdamax\int_{\region}dx)(\lambdamax)^{M+N} \prod_{n=1}^{N} \sigma(\f(\x_n))\prod_{m=1}^{M} \sigma(-\f(\x_m)). %\numberthis
	\label{eq:tuples_extended_lik}
\end{align}
There is a crucial difference between \eq \ref{eq:tuples_extended_lik} and the usual likelihood considered in \adams. \eq \ref{eq:tuples_extended_lik} represents a distribution over tuples and thus, as mentioned above, is properly normalized. In addition, it makes a distinction between the observed and latent events and it is thus different from \eq \ref{eq:intractable_likelihood} written for the the tuple $(M+N, \{\x_1, ..., \x_N, \y_1, ..., \y_M\})$.
We can write the full joint distribution as $\mathcal{L}^{+}_{N+M} \defeq \mathcal{L}(\{\x_n\}_{n=1}^{N}, \{\y_m\}_{m=1}^{M}, M, \f, \lambdamax| \tau, \hyperparam)$:
\begin{equation}
\label{eq:joint_lik}
 \mathcal{L}^{+}_{N+M} = \frac{(\lambdamax)^{N+M}\exp(-\lambdamax\int_{\region}dx) }{N!M!}
\prod_{n=1}^{N} \sigma(\f(\x_n))  \prod_{m=1}^{M} \sigma(-\f(\y_m)) \times p(\f) \times p(\lambdamax)  ,
\end{equation}
where $p(\f) \defeq  p(\f_{N+M}) $ denotes the joint prior at both  $\{\x_n\}_{n=1}^N$ and  $\{\y_m\}_{m=1}^M$ and $p(\lambdamax)$ denotes the prior over the upper bound of the intensity function. We consider $p(\lambdamax) = \text{Gamma}(a, b)$ and set $a$ and $b$ so that $\lambdamax$ as has mean and standard deviation equal to 2$\times$ and 1$\times$ the intensity we would expect  from an homogenous Poisson process on $\region$.

\subsection{Scalability via inducing variables}  
As in standard \gptext modulated models, the introduction of a \gptext prior poses significant computational challenges during posterior estimation as inference would be dominated by algebraic operations that are cubic on the number of observations.  In order to make inference scalable, we follow the inducing-variable approach proposed
by \cite{2009variational} and further developed by \cite{bonilla2016generic}. To this end, we consider an augmented prior $p(\f, \u)$ with $K$ underlying inducing variables denoted by $\u$.  The corresponding inducing inputs are given by the $K \times D$ matrix $\Z$. Major computational gains are realized when $K \ll N+M$. % |\{\x_n,\y_m\}_{n=1,m=1}^{N, M}|$. 
The augmented prior distributions for the inducing variables and the latent functions are $p(\u | \hyperparam) = \normal(\vec{0}, \Kzz) $ and $p( \f| \u, \hyperparam) = \normal(\Kxz  \Kzzinv \u,  \Kv - \A \Kzx) $ where $\A = \Kxz \Kzzinv$. 
The  matrices $\Kv$, $\Kxz$, $\Kzx$ and $\Kzzinv$ are the covariances induced by evaluating the corresponding covariance functions at all pairwise rows of the event locations $\{\x_n,\y_m\}_{n=1,m=1}^{N, M}$ and the inducing inputs $\Z$.
 \begin{figure} 
 	\centering
 	\begin{tikzpicture}[shorten >=1pt, node distance=2.0cm, on grid, auto,thick,scale=0.3]
 
 	\tikzstyle{main}=[circle, minimum size = 8mm, thick, draw =black!80, node distance = 17mm]
 	\tikzstyle{observed}=[circle, minimum size = 8mm, thick, draw =black!80, fill = black!10, node distance = 17mm]
 	\tikzstyle{constant}=[rectangle, minimum size = 6mm, thick, draw =black!80, node distance = 14mm, inner sep=0.1mm]
 	\tikzstyle{connect}=[-latex, thick]
 	
 	\node[main] (lambda_max) {$\lambdamax$};
 	\node[constant] (alpha) [above left=0.6cm and 1.5cm of lambda_max, label=below:$$] {$\alpha$};
 	\node[constant] (beta) [below left=0.6cm and 1.5cm of lambda_max, label=below:$$] {$\beta$};
 	
 	\node[main] (counts_M) [right= of lambda_max,label=below:$$] {$M$};
 	\node[main] (location_M) [right= of counts_M,label=below:$$] {$\y_m$};

 	\node[main] (inducing_process) [below= of counts_M,label=below:$$] {$u$};
 	\node[constant] (theta) [above left=0.6cm and 1.5cm of inducing_process, label=below:$$] {$\hyperparam$};
 	\node[observed] (inducing_inputs) [below left=0.6cm and 1.5cm of inducing_process,label=below:$$] {$Z_{d}$};
 	\node[main] (latent_process) [right=of inducing_process,label=below:$$] {$f$};

 	%\node[observed] (counts_N) [right= of latent_process,label=below:$$] {$N$};
 	%\node[observed] (location_N) [right= of counts_N,label=below:$$] {$\x_n$};
 
 	\node[rectangle, inner sep=0mm, fit= (location_M) ,label=below right: \small M, xshift=-1.5mm, yshift=1.2mm] {};
 	\node[rectangle, inner xsep=2.8mm, inner ysep=2.8mm,draw=black!100, fit = (location_M) ] (rect) {};
  	
  	%\node[rectangle, inner sep=0mm, fit= (location_N) ,label=below right:\small N, xshift=-1.5mm, yshift=1.2mm] {};
 	%\node[rectangle, inner xsep=2.8mm, inner ysep=2.8mm,draw=black!100, fit = (location_N) ] (rect) {};
 
  	\node[rectangle, inner sep=0mm, fit= (latent_process) ,label=below left:\small N+M, xshift=6.mm, yshift=1.2mm] {};
 	\node[rectangle, inner xsep=2.8mm, inner ysep=2.8mm,draw=black!100, fit = (latent_process) ] (rect) {};
 
   	\node[rectangle, inner sep=0mm, fit= (inducing_process) , label=below left:\small K, xshift=1.2mm, yshift=1.2mm] {};
 	\node[rectangle, inner xsep=2.8mm, inner ysep=2.8mm,draw=black!100, fit = (inducing_process) ] (rect) {};

 	\path (theta) edge [connect] (inducing_process)
 	(inducing_process) edge [connect] (latent_process)
 	(alpha) edge [connect] (lambda_max)
 	(beta) edge [connect] (lambda_max)
 	%(lambda_max) edge [connect] (counts_N)
 	(lambda_max) edge [connect] (counts_M)
 	
 	%(latent_process) edge [connect] (counts_N)
 	(inducing_process) edge [connect] (counts_M)
 	(latent_process) edge [connect, dashed] (location_M)
 	(counts_M) edge [connect] (location_M)
 	
 	%(counts_N) edge [connect] (location_N)
 	%(latent_process) edge [connect, bend right=50] (location_N)

 	(inducing_inputs) edge [connect] (inducing_process);
 	
 	\end{tikzpicture}
 	\caption{Plate diagram representing the posterior distribution accounting for all model dependencies. 
 		The only factorisation we introduce in our variational posterior (\eq \eqref{eq:variational_posterior}) is given by the dashed line.
 	} 
 	\label{fig:plates}
 \end{figure}
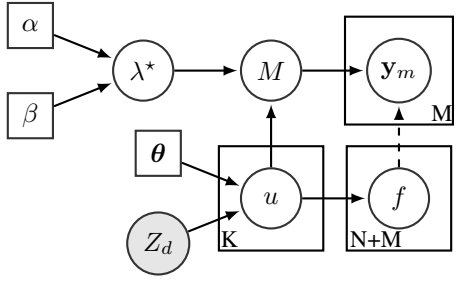

%\begin{center}
%	\begin{figure}[h]
%		\centering
%		\includegraphics[width=0.6\textwidth,height=\textheight,keepaspectratio]{images/plate}
%		\caption{\our - Plate diagram showing the posterior distribution accounting for all model dependencies. 
%			the dependency structure of our model. The dashed edge represents the only factorisation assumed in the variational distribution \eq \eqref{eq:variational_posterior}.}
%	\end{figure}
%	\label{fig:plates}
%\end{center}

\section{Structured Variational Inference in the augmented space}
Given the joint distribution in \eq \ref{eq:joint_lik}, our goal is to estimate the posterior distribution over all latent variables given the data. \ie $p(\f,\u,M, \{\y_m\}_{m=1}^{M},\lambdamax|\mathcal{D})$. This posterior is analytically intractable and we resort to variational inference \cite{jordan1999introduction}. Variational  inference entails defining an approximate posterior $q(\f,\u,M, \{\y_m\}_{m=1}^{M},\lambdamax)$  and optimizing the so-called evidence lower bound (\termelbo) with respect to this distribution. % We thus minimize the Kullback-Leibler (\termkl) divergence  between the approximate posterior and the true posterior. 
In \sgcp, the \gptext and the latent variables are highly coupled and breaking their dependencies would lead to poor approximations, especially in high dimensional settings. \fig \ref{fig:plates} shows the structure of a general posterior distribution for \sgcp without any factorisation assumption. We consider an approximate posterior distribution that takes dependencies into account:
\begin{align}
Q(\f, \u, M, \{\y_m\}_{m=1}^{M},\lambdamax)  = p(\f|\u) q(\{\y_m\}_{m=1}^{M}|M) q(M|\u, \lambdamax) q(\u)q(\lambdamax)
\label{eq:variational_posterior}
\end{align}

With respect to the general posterior distribution, the only factorisation we impose in \eq \eqref{eq:variational_posterior} is in the factor $q(\{\y_m\}_{m=1}^{M}|M)$ where we drop the dependency on $\f$, see dashed line in \fig \ref{fig:plates}. We set:
\begin{align*}
	& q(\u) = \mathcal{N}(\varmeanu, \varcovu) \quad  \quad q(\lambdamax) = \text{Gamma}(\alpha, \beta) \quad q(\{\y_m\}_{m=1}^{M}|M) = \prod_{m=1}^{M}\sum_{s=1}^{S}\pi_s \mathcal{N}_T(\mu_s, \sigma^2_s; \region)
\end{align*}
 where $\mathcal{N}_T(\cdot;\region)$ denotes a truncated Gaussian distribution on $\region$. 
More importantly, we assume $q(M|\u, \lambdamax) = \text{Poisson}(\eta)$ with $\eta = \lambdamax\int_{\region}\sigma(-\u(\x))d\x$. This is indeed the \emph{true} conditional posterior distribution for the number of thinned points, see Proposition (3.7) in \cite{moller2003statistical}. Considering $q(M|\u, \lambdamax)$ we thus fully account for the dependency structure existing among $M$, $\u$ and $\lambdamax$. 
Crucially, our algorithm does not require \emph{accurate} estimation of $\int_{\region}\sigma(-\u(\x))d\x$.  Differently from the competing techniques, where the algorithm convergence and the posterior $q(\f)$ is \emph{directly} dependent on numerical integration, we only require evaluation of the integral during the optimisation but $q(\f)$ and thus $\lambda(\x)$ do not directly depend on its value. 
%A very low number of point is sufficient to get good approximations for $q(M|\u, \lambdamax)$.

\subsection{Evidence Lower Bound}
Following standard variational inference arguments, it is straightforward to show that the \termelbo decomposes as:
\begin{align*}
\elbo
%&\expectation{Q}{\log\left[\frac{p(\{\x_n\}_{n=1}^{N}, \{\y_m\}_{m=1}^{M}, M, \f, \u, \lambdamax| \region, \hyperparam)}{p(\f|\u) q(\{\y_m\}_{m=1}^{M}|M) q(M|\u, \lambdamax) q(\u)q(\lambdamax)}\right]} 
& = N (\psi(\alpha) - \log(\beta)) - V\frac{\alpha}{\beta} - \log(N!) +  \underbrace{\expectation{Q}{M \log(\lambdamax)}}_{T_1}   -  \underbrace{\expectation{Q}{\log(M!)}}_{T_2} + 
\\& +\sum_{n=1}^{N} \expectation{q(\f)}{\log(\sigma(\f(\x_n)))} 
+ \underbrace{\expectation{Q} {\sum_{m=1}^{M} \log(\sigma(-\f(\y_m)))}}_{T_3} - \klterm^{\u}- \klterm^{\lambdamax} -  \underbrace{\enterm^M}_{T_4}- \underbrace{\enterm^{\{\y_m\}_{m=1}^M}}_{T_5} \numberthis
\label{eq:elbo}
%\expectation{Q}{\log q(M|\u, \lambdamax)} -  \underbrace{\expectation{Q}{\sum_{m=1}^{M}\log q(\y_m)}}
\end{align*}
where $V = \int_{\region}d\x$, $\psi(\cdot)$ is the digamma function and $q(\f) = \mathcal{N}(\A\varmeanu,\Kv - \A\Kzx +\A \varcovu \A')$ . The terms denoted by $T_i, i=1,...,5$ cannot be computed analytically. Na\"{i}vely, black-box variational inference algorithms could be used to estimate these terms via Monte Carlo, thus sampling from the full variational posterior (\eq \eqref{eq:variational_posterior}). This would require sampling $\f$, $\lambdamax$, $M$ and $\{\y_m\}_{m=1}^M$ thus slowing down the algorithm while leading to slow convergence. On the contrary, we exploit the structure of the model and the approximate posterior to simplify these terms and increase the algorithm efficiency. Denote $\parameterM = \int_{\region}{\sigma(-\u(\x))d\x}$, we can write:

\begin{align}
	&T_1 = \expectation{q(\lambdamax)}{\lambdamax\log(\lambdamax)} \expectation{q(\u)}{\parameterM}
	\quad  T_2 \leq \left[\expectation{q(\lambdamax)}{\lambdamax\log(\lambdamax)} + \frac{\alpha}{\beta}\right]\expectation{q(\u)}{\parameterM\log(\parameterM)} \label{eq:derivations1} \\
	&T_3  = \frac{\alpha}{\beta}\expectation{q(\u)}{\parameterM}\expectation{q(\f)q(\y_m)}{\log(\sigma(-\f(\y_m)))} \quad \quad T_5 =\frac{\alpha}{\beta} \expectation{q(\y_m)}{\log q(\y_m)} \expectation{q(\u)}{\parameterM} \label{eq:derivations2}\\
	&T_4 \leq \frac{\alpha}{\beta}\expectation{q(\u)}{\parameterM(\log(\parameterM)-1)} + \expectation{q(\lambdamax)q(\u)}{\lambdamax\log(\lambdamax)\expectation{q(\u)}{\parameterM-1}} 
	\label{eq:derivations}
\end{align}
where we use Stirling's approximation for $\log M! \approx M\log(M) - M$ and the bound $\eta \log(\eta) \leq \expectation{}{M\log(M)} \leq \eta \log(\eta + 1)$. See the supplement (\S1) for the full derivations.  \eqs \eqref{eq:derivations1}--\eqref{eq:derivations} give a lower bound for $\elbo$ which avoids sampling from $ q(M|\u, \lambdamax)$ and $q(\{\y_m\}_{m=1}^{M}|M)$ and does not require computing the \gptext on the stochastic locations. The remaining expectations are with respect to reparameterizable distributions. We thus avoid the use of score function estimators which would lead to high-variance gradient estimates. Stochastic optimisation techniques can be used to evaluate $T_3$ and $\sum_{n=1}^{N} \expectation{q(\f)}{\log(\sigma(\f(\x_n)))}$ making the corresponding computational cost independent of $M$ and $N$.  This would reduce the computational complexity of the algorithm to $\mathcal{O}(K^3)$. However, when the number of inputs used per mini-batch equals $N$, the time complexity becomes $\mathcal{O}(NK^2)$. In the following experiments, we show how the proposed structured approach together with these efficient \termelbo computations lead to higher predictive performances and better uncertainty quantification. 
%We exploit the described advantages in terms of structured posterior and computational complexity to achieve better predictive performances and uncertinty quantification iin the following experiments. 
%We will show in the following experiments how the structured approach together with the computationally efficient \termelbo computations lead to higher predictive performances and better uncertainty quantification. 
%\input{mt_extension}
\section{Experiments}
We test our algorithm on three 1\acro{d} synthetic data settings and on two 2\acro{d} real-world applications\footnote{Code and data for all the experiments will be provided.}. 

\textbf{Baselines}
We compare against alternative inference schemes, different link functions and a different augmentation scheme. In terms of continuous models, we consider the sampling approach of \citet{adams2009tractable}(\adams), a Permanental Point process model  \cite{lloyd-icml-2015}(\lloyds) and a mean-field  approximation based on a  P\'{o}lya-Gamma augmentation \cite{donner2018efficient}(\donner). In addition, we compare against a discrete variational log Gaussian Cox process model \cite{nguyen2014automated}(\lgcp). Details are given in the supplement (\S3).
 
\textbf{Performance measures}
We test the algorithms evaluating the $\lnorm$ norm to the true intensity function (for the synthetic datasets), the test log likelihood ($\ltest$) on the test set and the negative log predicted likelihood (\nlpl) on the training set. In order to assess the model capabilities in terms of uncertainty quantification, we compute the empirical coverage (\coverage), \ie the coverage of the empirical count distributions obtained by sampling from the posterior intensity. We do that for different credible intervals (\credible)  on both the training (in-sample, $p(N|\mathcal{D})$) and test set (out-of-sample, $p(N^*|\mathcal{D})$). Details on the metrics are in the supplement (\S2).
For the synthetic data experiments, we run the algorithms with 10 training datasets each including a different \ppp realization sampled from the ground truth.  For each different training set, we then evaluate the performance on other 10 unseen realizations sampled again from the ground truth. We compute the mean and the standard deviation for the presented metrics averaging across the training and test sets. For the real data settings, we compute the \nlpl and in-sample \coverage on the observed events. We then test the algorithm computing both $\ltest$ and out-of-sample \coverage on the held-out events. In order to compute the out-of-sample \coverage we rescale the intensity function as $\lambda_{test}(\x)=  \lambda_{train}(\x) - N_{train}/V + N_{test}/V$ with $V = \int_{\region}d\x$. We then sample from $\lambda_{test}(\x)$ and generate the predicted count distributions for different seeds. 
\begin{figure}
		\centering
		\includegraphics[width=0.8\textwidth,height=\textheight,keepaspectratio]{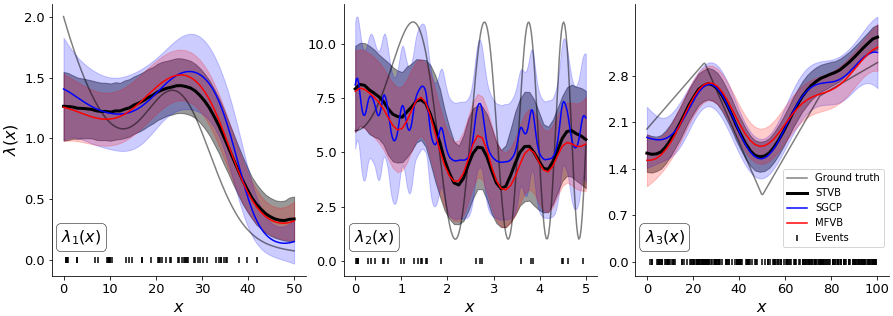}
		\caption{Qualitative results on synthetic data. Solid colored lines denote posterior mean intensities while shaded areas are $\pm$ standard deviation.}
		\label{fig:new_synt}
\end{figure}

%in-sample \coverage gives the number of count distributions containing the true observed number of events for different credible intervals (\credible). 
%The out-of-sample \coverage for the synthetic data is computed looking at the number of events in 10 unseen realisations. For the real data, it is computed by rescaling the intensity functions given the number of unseen events and generating the predicted count distributions for different seeds. 
\textbf{Synthetic experiments}
We test our approach using the three toy example proposed by \cite{adams2009tractable}:
\begin{itemize}
	\item $\lambda_1(\x) = 2\exp(-1/15) + \exp(-[(x-15)/10]^2)$ \quad  $x \in [0,50]$,
	\item $\lambda_2(\x) = 5\text{sin}(x^2) + 6$ \quad  $x \in [0,5]$ and
	\item $\lambda_3(\x)$ piecewise linear through $(0,20)$, $(25,3)$,$(50,1)$, $(75,2.5)$, $(100,3)$ \quad  $x \in [0,100]$.
\end{itemize}
For \lgcp, we discretize the input space considering a grid cell width of one for $\lambda_1(\x)$ and $\lambda_3(\x)$ and of 0.5 for  $\lambda_2(\x)$.  For \donner we consider 1000 integration points. In terms of $q(\{\y_m\}_{m=1}^M|M)$,  we set $S=5$ but consistent results where found across different values of this parameter. 
% results analysis
The results are given in \fig \ref{fig:new_synt} and \tbl \ref{tab:synthetic_1}, where we see that all algorithms recover similar predicted mean intensities and give roughly comparable performances across all metrics.  Out of all 9 settings and metrics (top section of \tbl \ref{tab:synthetic_1})  our method (\our) outperforms competing methods on 3 cases and it is only second to \adams on 6 cases. However, the \textsc{cpu} time  of \adams is almost an order of magnitude larger than ours even in these simple low-dimensional problems, making that method inapplicable to large datasets. This confirms the benefits of having structured approximate posteriors within  a computationally efficient inference algorithm such as \vi. In terms of uncertainty quantification (bottom section of \tbl \ref{tab:synthetic_1}), our algorithm outperforms  all competing approaches for $\lambda_1(\x)$ and  $\lambda_2(\x)$. %The discrete \lgcp has higher empirical coverage for $\lambda_3(\x)$. This is justified by the pattern observed in the events which is not zero-inflated making it easier for a discrete model to pick the significantly smoother function compared to the alternative settings.

\begin{table}
	\centering
	\caption{
	%\textit{Upper:} \crime dataset. Performance on the missing regions. Time in seconds per epoch. Lower values of \nlpl are better. \textit{Lower:} In-sample0.00Out-of-sample 90\% \ci coverage for the predicted event counts distributions. Higher values of \coverage are better.
% Synthetic data. Performances across the 10 training datasets and the 10 test sets. \textit{Lower:} Out-of-sample \coverage for different \credible. Standard errors in parenthesis. 
 % EB: REMEMBER NIPS PEOPLE ARE PRETY ANNOYING ABOUT DESCRIPTIVE CAPTIONS. THEY HAVE TO BE !
 Average performances on  synthetic data across 10 training and 10 test datasets with standard errors in brackets. Top: Lower values of $\lnorm$, $\nlpl$ and higher values of $\ltest$ are better. Bottom: Out-of-sample \coverage for different \credible, higher values are better. Our method denoted by \our. 
} 
	\begin{subtable}{0.8\textwidth}
		%\sisetup{table-format=-1.2}   % 2 decimals, leave space for minus sign
		\begin{center}
		\resizebox{\columnwidth}{!}{\begin{tabular}{@{\extracolsep{4pt}}@{\kern\tabcolsep}lcccccccccccccccccccccccccccccc}
				%{llllllrrrrrrrrrr}
				\toprule
				&\multicolumn{3}{c}{$\lambda_1(x)$} & \multicolumn{3}{c}{$\lambda_2(x)$} & \multicolumn{3}{c}{$\lambda_3(x)$} &	 \multirow{2}{*}{\cpu}\\
				\cmidrule(lr){2-4}
				\cmidrule(lr){5-7}
				\cmidrule(lr){8-10}
				& $\lnorm$   & $\ltest$     & \nlpl    &   $\lnorm$ &    $\ltest$  &   \nlpl &     $\lnorm$ &    $\ltest$  &   \nlpl  & time (s) \\
				\cmidrule(lr){2-4}
				\cmidrule(lr){5-7}
				\cmidrule(lr){8-10}
				\multirow{2}{*}{\our}      	&   \textbf{3.44}            &  \textbf{-1.39} & 4.71      			& 46.28           & 56.04    & 5.62				& \textbf{7.39}            & 153.98   & 6.41   					& \multirow{2}{*}{315.59}   \\
				& (1.43)          & (1.05) & (0.51)    		& (9.95)          & (4.47)  & (0.72)					& (2.76)          & (11.91)  & (0.64) 						&      \\
				\multirow{2}{*}{\donner}   & 4.56            & -2.84  & 4.74        	& 44.44           & 55.35   & 5.52  					& 8.17            &  155.08   & 5.82  							& \multirow{2}{*}{0.01}      \\
				& (1.43)          & (1.0)  & (0.1) 					& (10.7)          & (4.72)   & (1.29)					 & (3.43)          & (10.20)  & (0.61) 							&      \\
				\multirow{2}{*}{\lloyds}   & 9.19            & -7.71 & 8.91  				& 48.15           & \textbf{56.82}   & 5.20  					& 20.54           & 152.82   & 8.35  							& \multirow{2}{*}{0.44}     \\
				& (2.32)          & (3.31) & (1.19) 			& (13.16)         & (4.42)  & (1.33) 					& (6.53)          & (11.43)    & (2.28) 							&    \\
				\multirow{2}{*}{\adams}    & 4.22            & \textbf{-1.39}    & \textbf{4.21}	& \textbf{43.50}           & 55.05    & \textbf{3.77} 						& 14.44           &  \textbf{165.66}   & \textbf{4.78}  						& \multirow{2}{*}{2764.88}   \\
				& (1.88)          & (1.28) & (1.04) 			& (8.69)          & (1.35)  & (0.54) 					& (2.97)          & (2.12)   & (0.33) 						&       \\
				\multirow{2}{*}{\lgcp}     & 67.76           & -5.26 & 26.26 			& 106.74          & 28.56    & 15.75 					& 19.24           & 147.67    & 10.84 						& \multirow{2}{*}{4.74}      \\
				& (24.38)         & (8.84)  & (8.09) 		& (13.89)         & (6.88)   & (3.36) 					& (6.44)          & (11.76)   & (1.36) 						&       \\
				\bottomrule
		\end{tabular}}  
		\end{center}
	\end{subtable}
	
	\begin{subtable}{0.8\textwidth}
		\centering
		%\sisetup{table-format=4.0} % integer values only, up to 4 digits
		\begin{center}
		\resizebox{\columnwidth}{!}{\begin{tabular}{lcccccccccccccccccccccccccccccccccccc}
				%\hline
				& \multicolumn{3}{c}{\coverage -- $\lambda_1(x)$}  & \multicolumn{3}{c}{\coverage -- $\lambda_2(x)$}  & \multicolumn{3}{c}{\coverage -- $\lambda_3(x)$}  \\
				\cmidrule(lr){2-4}
				\cmidrule(lr){5-7}
				\cmidrule(lr){8-10}
				&  30\%   \credible                & 40\%        \credible          & 50\% \credible &  30\%   \credible                & 40\%    \credible              & 50\% \credible & 30\%       \credible            & 40\%        \credible          & 50\%      \credible            \\
				\cmidrule(lr){2-4}
				\cmidrule(lr){5-7}
				\cmidrule(lr){8-10}
				\our        & \textbf{0.81}                 & \textbf{0.72}                 & \textbf{0.6}           & \textbf{0.91}                 & \textbf{0.88}                 & \textbf{0.86}    &  \textbf{0.99}                & 0.97                 & 0.92             \\
				& (0.27)               & (0.27)               & (0.34)     & (0.24)               & (0.23)               & (0.22)     & (0.03)                & (0.09)                & (0.15)       \\

				\donner &  0.76                 & 0.61                 & 0.52              & 0.89                 & 0.84                 & 0.82       & 0.97                 & 0.91                 & 0.78      \\
				& (0.25)                & (0.28)                & (0.29)   & (0.23) & (0.29) & (0.29)      & (0.09)                & (0.14)                & (0.15)  \\

				\lloyds & 0.75                 & 0.41                 & 0.04             & 0.76                 & 0.45                 & 0.05        & 0.83                 & 0.43                 & 0.03   \\
				& (0.21)               & (0.25)               & (0.09)         & (0.26)               & (0.26)               & (0.05)   & (0.19)               & (0.14)               & (0.05)   \\

				\adams              & 0.39                 & 0.27                 & 0.08           & 0.64                 & 0.14                 & 0.00       & 0.49                 & 0.34                 & 0.02        \\
				& (0.28)               & (0.22)               & (0.12)         & (0.09)               & (0.05)               &  (0.00)  & (0.03)               & (0.07)               & (0.04)         \\

				\lgcp              & 0.08                 & 0.03                 & 0.01           & 0.04                 & 0.00               & 0.00      & \textbf{0.99}                  & \textbf{0.99}                  & \textbf{0.95}         \\
				& (0.12)                & (0.09)                & (0.03)                & (0.08)               & (0.00)                &  (0.00)         &  (0.00)                & (0.12)                & (0.10)             \\
				\bottomrule
		\end{tabular}}
		\end{center}
	\end{subtable}
	\label{tab:synthetic_1}
\end{table}

\begin{figure}[h]
		\centering
		\includegraphics[width=0.49\textwidth,height=\textheight,keepaspectratio]{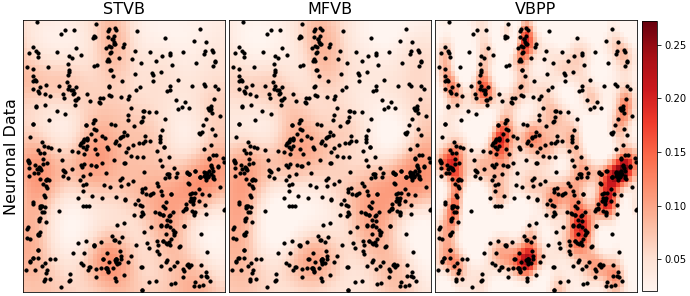}
	\includegraphics[width=0.48\textwidth,height=\textheight,keepaspectratio]{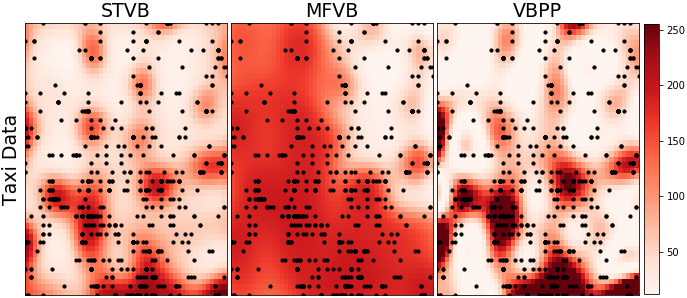}
	\caption{Real data. Posterior mean intensities and events on the two-dimensional input space.}
	\label{fig:real_intensity}
	\end{figure}

\begin{figure}[h]
		\centering
		\includegraphics[width=0.38\textwidth,height=4cm,]{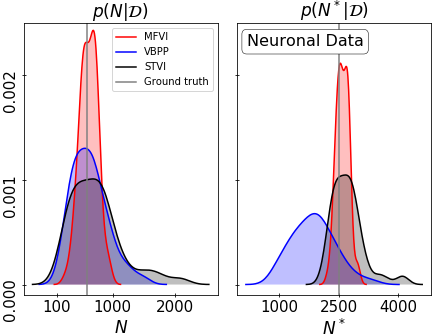}
		\includegraphics[width=0.49\textwidth,height=4cm]{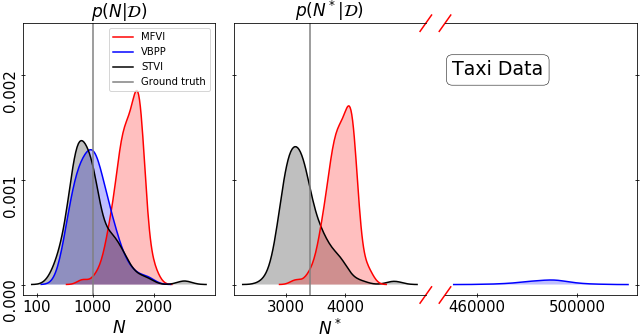}
		\caption{Predicted counts distributions for the training set ($p(N|\mathcal{D})$) and the test set ($p(N^*|\mathcal{D})$) on real data. The \textcolor{gray}{gray} line denotes the number of observed events. The \textcolor{red}{red} bars on the x-axis denote breaks in the axis due to the different shifts of the distributions. }
			\label{fig:real_hist}
	\end{figure}

\paragraph{2\acro{D} real data experiments}
In this section we show the performance of the algorithm on two 2\acro{D} real-world datasets. In both cases, we assume independent two-dimensional truncated Gaussian distributions for $q(\{\y_m\}_{m=1}^{M}|M)$ so that they factorize  across  input  dimensions.  Qualitative and quantitative results are given in \fig \ref{fig:real_intensity}, \fig \ref{fig:real_hist} and \tbl \ref{tab:performance_real}. 

Our first dataset is concerned with neuronal data, where event locations correspond to the position of a mouse moving in an arena when a recorded cell fired \cite{sargolini2006conjunctive, sargolini_data}. We randomly assign the events to either training ($N = 583$) or test  ($N = 29710$) and we run the model using a regular grid of $10 \times 10$  inducing inputs. We see that the intensity function recovered by the three methods vary in terms of smoothness with \donner estimating the smoothest $\lambda(\x)$ and \lloyds recovering an irregular surface (\fig \ref{fig:real_intensity}). \donner gives  slightly better performance in terms of $\ltest$ but our method (\our) outperforms competing approaches in terms of \nlpl and \coverage figures. Remarkably, \our contains the true number of test events in the 30\% credible intervals for 56\% of the simulations from the posterior intensity (\tbl \ref{tab:performance_real} and \fig \ref{fig:real_hist}).

As a second dataset, we consider the Porto taxi dataset\footnote{\texttt{http://www.geolink.pt/ecmlpkdd2015-challenge/dataset.html}} which contains the trajectories of 7000 taxi travels in the years 2013/2014 in the city of Porto. As in \cite{donner2018efficient}, we  consider the pick-up locations as observations of a \ppp and restrict the analysis to events happening within the coordinates $(41.147, -8.58)$ and $(41.18, -8.65)$. We select $N=1000$ events at random as training set and train the model with 400 inducing points placed on a regular grid. The test log likelihood is then computed on the remaining 3401 events. We see that our method (\our) outperforms competing methods on all performance metrics (\tbl \ref{tab:performance_real}), recovering an intensity that is smoother than \lloyds and captures more structure compared to \donner (\fig \ref{fig:real_intensity}). In terms of uncertainty quantification, the coverage of $p(N^*|\mathcal{D})$ are the highest for \our across all \credible. Notice how the irregularity of the \lloyds intensity leads to good performance on the training set but results in a $p(N^*|\mathcal{D})$ which is centered on a significantly higher number of test events (\fig \ref{fig:real_hist}). As expected, the \svi approach implies wider counts distributions compared to the mean field approximation. This generally yields better predictive performances in a variety of settings and especially in higher-dimensional experiments.  


\begin{table}
	\caption{Average performances on  real-data experiments with standard errors in brackets. \coverage is computed across 100 replications using different seeds. Higher $\ltest$, \coverage and lower \nlpl are better.   \coverage figures are given as In-sample - Out-of-sample.} 
	\begin{center}
	\resizebox{\columnwidth}{!}{\begin{tabular}{@{\extracolsep{4pt}}@{\kern\tabcolsep}lcccccccccccccccc}
		%{llllllrrrrrrrrrr}
		\toprule
		& \multicolumn{5}{c}{Neuronal data} & \multicolumn{5}{c}{Taxi data}\\
		\cmidrule(lr){2-6}
		\cmidrule(lr){7-11}
		  &  $\ltest [\times  10^3]$  &  \nlpl & \coverage -30\% \credible & \coverage -40\% \credible & \cpu time (s) &  $\ltest [\times  10^6]$ &  \nlpl $ [\times  10^3]$ & \coverage - 30\% \credible &\coverage - 40\% \credible  & \cpu time (s)\\
			\cmidrule(lr){2-6}
		\cmidrule(lr){7-11} 
		\multirow{2}{*}{\our}     &       -84.55 & \textbf{10.10} &   \textbf{1.00}-\textbf{1.00} & \textbf{0.99}-\textbf{0.56}  &  \multirow{2}{*}{ 193.07} &   \textbf{-27.96} & \textbf{27.96} & 0.81-\textbf{0.37}     & 0.09-\textbf{0.01} &  \multirow{2}{*}{290.34}\\
		& (16.05) & (7.02) & (0.00)-(0.00)& (0.10)-(0.50)  & & (9.16)& (91.58) & (0.39)-(0.48) & (0.29)-(0.10) \\
		\multirow{2}{*}{\donner}   &      \textbf{ -83.54} & 10.71 &  \textbf{1.00}-0.03 & 0.78-0.00 &    \multirow{2}{*}{0.35} &   -40.8 & 40.65 &  0.00-0.00        &          0.00-0.00   &   \multirow{2}{*}{0.24}\\
			& (4.60)& (3.39) & (0.00)-(0.17) & (0.41)-(0.00)  & & (6.41) & (64.14)  & (0.00)-(0.00)        &          (0.00)-(0.00) \\
		\multirow{2}{*}{\lloyds}   &       -83.89 & 11.39 &  \textbf{1.00}-0.00 & 0.83-0.00 &    \multirow{2}{*}{26.23} & -31.32 &  31.32 & \textbf{0.98}-0.00     & \textbf{0.48}-0.00   &\multirow{2}{*}{3.62}\\
			& (12.49) & (8.18) & (0.00) - (0.00)& (0.38)-(0.00) & &(8.18) & (81.83) & (0.14)-(0.00)    & (0.50)-(0.00)\\
		\bottomrule
	\end{tabular}}
\label{tab:performance_real}
\end{center}
\end{table}

\section{Conclusions and discussion}
We have proposed a new variational inference framework for estimating the intensity of a continuous sigmoidal Cox process. By seeing an augmented input space from a superposition of two \ppp{s},  we have derived a scalable and computationally efficient structured variational approximation. Our framework does not require discretization or accurate numerical computation of integrals on the input space, it is not limited to specific kernel functions and properly accounts for the strong dependencies existing across the latent variables. 
Through extensive empirical evaluation we have shown that our methods compares favorably  against `exact'  but computationally costly \mcmc schemes, while being almost an order of magnitude faster. More importantly, our inference scheme outperforms all competing approaches in terms of uncertainty quantification. The benefit of the proposed scheme and resulting \svi are particularity pronounced on multivariate input settings where accounting for the highly coupled variables become crucial for interpolation and prediction.  
Future work will focus on relaxing the factorization assumption between the \gptext and the latent points. Introducing a fully structured variational inference would further improve the accuracy performance of the method but would require further approximations in the variational objective.
%gwhich could increase the computational cost. 

\bibliographystyle{apalike}
\bibliography{references}

\begin{thebibliography}{}

\bibitem[Adams et~al., 2009]{adams2009tractable}
Adams, R.~P., Murray, I., and MacKay, D.~J. (2009).
\newblock Tractable nonparametric bayesian inference in poisson processes with
  gaussian process intensities.
\newblock In {\em Proceedings of the 26th Annual International Conference on
  Machine Learning}, pages 9--16. ACM.

\bibitem[Aglietti et~al., 2018]{aglietti2018}
Aglietti, V., Damoulas, T., and Bonilla, E. (2018).
\newblock Log gaussian cox process networks.
\newblock {\em arXiv preprint arXiv:1805.09781}.

\bibitem[Bonilla et~al., 2016]{bonilla2016generic}
Bonilla, E.~V., Krauth, K., and Dezfouli, A. (2016).
\newblock Generic inference in latent {G}aussian process models.
\newblock {\em arXiv preprint arXiv:1609.00577}.

\bibitem[Brix and Diggle, 2001]{brix2001spatiotemporal}
Brix, A. and Diggle, P.~J. (2001).
\newblock Spatiotemporal prediction for log-gaussian cox processes.
\newblock {\em Journal of the Royal Statistical Society: Series B (Statistical
  Methodology)}, 63(4):823--841.

\bibitem[cell data Sargolini et~al 2006.https://doi.org/10.11582/2014.00003,
  2014]{sargolini_data}
cell data Sargolini et~al 2006.https://doi.org/10.11582/2014.00003, G. (2014).
\newblock Centre for the biology of memory and francesca sargolini.

\bibitem[Cox, 1955]{cox1955some}
Cox, D.~R. (1955).
\newblock Some statistical methods connected with series of events.
\newblock {\em Journal of the Royal Statistical Society. Series B
  (Methodological)}, pages 129--164.

\bibitem[Cunningham et~al., 2008]{cunningham2008fast}
Cunningham, J.~P., Shenoy, K.~V., and Sahani, M. (2008).
\newblock Fast gaussian process methods for point process intensity estimation.
\newblock In {\em Proceedings of the 25th international conference on Machine
  learning}, pages 192--199. ACM.

\bibitem[Daley and Vere-Jones, 2003]{daley2007introduction}
Daley, D.~J. and Vere-Jones, D. (2003).
\newblock {\em An Introduction to the Theory of Point Processes. Volume I:
  Elementary Theory and Methods}.
\newblock Springer Science \& Business Media.

\bibitem[Diggle et~al., 2013]{diggle2013spatial}
Diggle, P.~J., Moraga, P., Rowlingson, B., and Taylor, B.~M. (2013).
\newblock Spatial and spatio-temporal log-gaussian cox processes: Extending the
  geostatistical paradigm.
\newblock {\em Statistical Science}, pages 542--563.

\bibitem[Donner and Opper, 2018]{donner2018efficient}
Donner, C. and Opper, M. (2018).
\newblock Efficient bayesian inference of sigmoidal gaussian cox processes.
\newblock {\em The Journal of Machine Learning Research}, 19(1):2710--2743.

\bibitem[Grubesic and Mack, 2008]{grubesic2008spatio}
Grubesic, T.~H. and Mack, E.~A. (2008).
\newblock Spatio-temporal interaction of urban crime.
\newblock {\em Journal of Quantitative Criminology}, 24(3):285--306.

\bibitem[Gunter et~al., 2014]{gunter2014efficient}
Gunter, T., Lloyd, C., Osborne, M.~A., and Roberts, S.~J. (2014).
\newblock Efficient bayesian nonparametric modelling of structured point
  processes.
\newblock {\em arXiv preprint arXiv:1407.6949}.

\bibitem[John and Hensman, 2018]{john2018large}
John, S. and Hensman, J. (2018).
\newblock Large-scale cox process inference using variational fourier features.
\newblock {\em arXiv preprint arXiv:1804.01016}.

\bibitem[Jordan et~al., 1999]{jordan1999introduction}
Jordan, M.~I., Ghahramani, Z., Jaakkola, T.~S., and Saul, L.~K. (1999).
\newblock An introduction to variational methods for graphical models.
\newblock {\em Machine learning}, 37(2):183--233.

\bibitem[Lasko, 2014]{lasko2014efficient}
Lasko, T.~A. (2014).
\newblock Efficient inference of gaussian-process-modulated renewal processes
  with application to medical event data.
\newblock In {\em Uncertainty in artificial intelligence: proceedings of the...
  conference. Conference on Uncertainty in Artificial Intelligence}, volume
  2014, page 469. NIH Public Access.

\bibitem[Lewis and Shedler, 1979]{lewis1979simulation}
Lewis, P.~W. and Shedler, G.~S. (1979).
\newblock Simulation of nonhomogeneous poisson processes by thinning.
\newblock {\em Naval research logistics quarterly}, 26(3):403--413.

\bibitem[Lian et~al., 2015]{pmlr-v37-lian15}
Lian, W., Henao, R., Rao, V., Lucas, J., and Carin, L. (2015).
\newblock A multitask point process predictive model.
\newblock In {\em International Conference on Machine Learning}, pages
  2030--2038.

\bibitem[Lloyd et~al., 2016]{lloyd2016latent}
Lloyd, C., Gunter, T., Nickson, T., Osborne, M., and Roberts, S.~J. (2016).
\newblock Latent point process allocation.

\bibitem[{Lloyd} et~al., 2015]{lloyd-icml-2015}
{Lloyd}, C., {Gunter}, T., {Osborne}, M.~A., and {Roberts}, S.~J. (2015).
\newblock {Variational Inference for Gaussian Process Modulated Poisson
  Processes}.
\newblock In {\em International Conference on Machine Learning}.

\bibitem[L{\'o}pez-Lopera et~al., 2019]{lopez2019gaussian}
L{\'o}pez-Lopera, A.~F., John, S., and Durrande, N. (2019).
\newblock Gaussian process modulated cox processes under linear inequality
  constraints.
\newblock {\em Artificial Intelligence and Statistics}.

\bibitem[Marsan and Lengline, 2008]{marsan2008extending}
Marsan, D. and Lengline, O. (2008).
\newblock Extending earthquakes' reach through cascading.
\newblock {\em Science}, 319(5866):1076--1079.

\bibitem[M{\o}ller et~al., 1998]{moller1998log}
M{\o}ller, J., Syversveen, A.~R., and Waagepetersen, R.~P. (1998).
\newblock Log gaussian cox processes.
\newblock {\em Scandinavian journal of statistics}, 25(3):451--482.

\bibitem[Moller and Waagepetersen, 2003]{moller2003statistical}
Moller, J. and Waagepetersen, R.~P. (2003).
\newblock {\em Statistical inference and simulation for spatial point
  processes}.
\newblock Chapman and Hall/CRC.

\bibitem[Nguyen and Bonilla, 2014]{nguyen2014automated}
Nguyen, T.~V. and Bonilla, E.~V. (2014).
\newblock Automated variational inference for {G}aussian process models.
\newblock pages 1404--1412.

\bibitem[Rasmussen and Williams, 2005]{Rasmussen:2005:GPM:1162254}
Rasmussen, C.~E. and Williams, C. K.~I. (2005).
\newblock {\em Gaussian Processes for Machine Learning}.
\newblock The MIT Press.

\bibitem[Sargolini et~al., 2006]{sargolini2006conjunctive}
Sargolini, F., Fyhn, M., Hafting, T., McNaughton, B.~L., Witter, M.~P., Moser,
  M.-B., and Moser, E.~I. (2006).
\newblock Conjunctive representation of position, direction, and velocity in
  entorhinal cortex.
\newblock {\em Science}, 312(5774):758--762.

\bibitem[Titsias, 2009]{2009variational}
Titsias, M.~K. (2009).
\newblock Variational learning of inducing variables in sparse gaussian
  processes.
\newblock 5:567--574.

\bibitem[Walder and Bishop, 2017]{pmlr-v70-walder17a}
Walder, C.~J. and Bishop, A.~N. (2017).
\newblock Fast {B}ayesian intensity estimation for the permanental process.
\newblock In {\em International Conference on Machine Learning}.

\end{thebibliography}

\end{document}

\maketitle

\section{ELBO derivations}
Here we derive the expressions given in \eqs (7)-(10). As given in \eq (7) the evidence lower bound ($\elbo$) decomposes as:

\begin{align*}
& \elbo = \expectation{Q}{\log\left[\frac{p(\{\x_n\}_{n=1}^{N}, \{\x_m\}_{m=1}^{M}, M, \f, \u, \lambdamax| \tau, \hyperparam)}{p(\f|\u) q(\{\x_m\}_{m=1}^{M}|M) q(M|\u, \lambdamax) q(\u)q(\lambdamax)}\right]} 
\\& = \expectation{Q}{\log{p(\{\x_n\}_{n=1}^{N}, \{\x_m\}_{m=1}^{M}, M, \u,\lambdamax| \tau, \hyperparam)}} 
- \expectation{Q}{\log q(\{\x_m\}_{m=1}^{M}|M) q(M|\u, \lambdamax) q(\u)q(\lambdamax)}
\\& =\mathbb{E}_Q((N + M)\log(\lambdamax) - \lambdamax \mu(\tau) - \log(M!) - \log(N!) + \sum_{n=1}^{N}\log(\sigma(\f(\x_n))) \\& +  \sum_{m=1}^{M}\log(\sigma(-\f(\x_m)) +  \log(p(\u))   +  \log(p(\lambdamax))
\\ & - \log(q(\u)) - \log(q(M|\u, \lambdamax)) - \log(q(\lambdamax)) - \log(q(\{\x_m\}_{m=1}^{M}|M))
\\& = N (\psi(\alpha)-\log(\beta)) - V\frac{\alpha}{\beta} - \log(\log N!) + \underbrace{\expectation{Q}{M \log(\lambdamax)}}_{T_1}  -  \underbrace{\expectation{Q}{\log M!}}_{T_2}  + 
\\& +\sum_{n=1}^{N} \expectation{q(\u)}{\log(\sigma(\f(\x_n)))} 
+ \underbrace{\expectation{Q} {\sum_{m=1}^{M} \log(\sigma(-\f(\x_m)))}}_{T_3}+
\\& - KL(q(\u)||p(\u)) - KL(q(\lambdamax)||p(\lambdamax)) 
\\& -  \underbrace{\expectation{Q}{\log q(M|\u, \lambdamax)}}_{T_4} -  \underbrace{\expectation{Q}{\log q(\{x_m\}_{m=1}^{M}|M)}}_{T_5}
\end{align*}

Let's now focus on the terms $T_i, i=1, ..., 5$. In the following derivations we will exploit the Stirling's approximation for $\log(M!)\approx M\log M - M$ and the bounds $\eta \log \eta \leq \expectation{Q}{M \log M}\leq \eta \log(\eta +1)$ to obtain a lower bound for the $\elbo$. The term $T_1$ (\eq (8)) is given by:
\begin{align*}
T_1 &= \expectation{q(\u)q(\lambdamax)}{\expectation{q(M|\u, \lambdamax)}{M\log(\lambdamax)}}
\\& = \expectation{q(\u)q(\lambdamax)}{\log(\lambdamax)  \expectation{q(M|\u, \lambdamax)}{M}}
\\& = \expectation{q(\u)q(\lambdamax)}{\log(\lambdamax)  \lambdamax \int_{\tau}\sigma(-\u(x))dx}
\\& = \expectation{q(\lambdamax)}{\lambdamax\log(\lambdamax)} \expectation{q(\u)}{\parameterM}  
\end{align*}

The term $T_2$ (\eq (8)) is given by:

\begin{align*}
T_2 & \approx \expectation{Q}{M\log(M)} - \expectation{Q}{M}
\\& = \expectation{Q}{M\log(M)} - \expectation{q(\lambdamax)q(\u)}{\expectation{q(M|\lambdamax, \u)}{M}}
\\& = \expectation{Q}{M\log(M)} - \expectation{q(\lambdamax)q(\u)}{\lambdamax  \int_{\tau}\sigma(-\u(x))dx}
\\& = \expectation{Q}{M\log(M)} - \expectation{q(\lambdamax)}{\lambdamax } \expectation{q(\u)}{\int_{\tau}\sigma(-\u(x))dx}
\\& = \expectation{Q}{M\log(M)} - \frac{\alpha}{\beta} \expectation{q(\u)}{\parameterM}
\\& \geq \expectation{q(\lambdamax)q(\u)}{\eta \log (\eta +1)} - \frac{\alpha}{\beta} \expectation{q(\u)}{\parameterM}
\\& = \expectation{q(\lambdamax)}{\lambdamax\log(\lambdamax)}\expectation{q(\u)}{\parameterM} + \frac{\alpha}{\beta}\expectation{q(\u)}{\parameterM\log(\parameterM)} - \frac{\alpha}{\beta} \expectation{q(\u)}{\parameterM}
\end{align*}

The term $T_3$ (\eq (9)) is given by:
\begin{align*}
T_3 &= \expectation{q(\f)q(\u)q(\y_m)q(\lambdamax)}{\expectation{q(M|\f, \y_m)}{\sum_{m=1}^{M}\log(\sigma(-\f(\y_m)}}
\\& =  \expectation{q(\f)q(\u)q(\y_m)q(\lambdamax)}{\log(\sigma(-\f(\y_m) \expectation{q(M|\f, \y_m)}{\sum_{m=1}^{M}1}}
\\& =  \expectation{q(\f)q(\u)q(\y_m)}{\log(\sigma(-\f(\y_m) \lambdamax \parameterM}
\\& =\frac{\alpha}{\beta}\expectation{q(\u)}{\parameterM}\expectation{q(\f)q(\y_m)}{\log(\sigma(-\f(\y_m)))}
\end{align*}

The term $T_4$ (\eq (10)) is given by:

\begin{align*}
T_4 &= \expectation{Q}{-\lambdamax\parameterM} + \expectation{Q}{M\log(\lambdamax\parameterM}  -\expectation{Q}{\log M!}
\\& = - \frac{\alpha}{\beta}\expectation{q(\u)}{\int_{\tau}\sigma(-\u(x))dx}
+ \expectation{q(\lambdamax)q(\u)}{\lambdamax\log(\lambdamax)\parameterM + \lambdamax\parameterM\log(\parameterM)}\expectation{q(\u)}{\int_{\tau}\sigma(-\u(x))dx}
\\& - \expectation{Q}{\log M!}
\\& = - \frac{\alpha}{\beta}\expectation{q(\u)}{\int_{\tau}\sigma(-\u(x))dx} +\expectation{q(\lambdamax)}{\lambdamax\log(\lambdamax)}\expectation{q(\u)}{\parameterM}
\\& + \frac{\alpha}{\beta}\expectation{q(\u)}{\parameterM\log(\parameterM)} 
\\& + \frac{\alpha}{\beta} \expectation{q(\u)}{\parameterM} - \expectation{Q}{M\log(M)}
\end{align*}

Substituting the bound for $\expectation{Q}{M\log(M)}$ and simplifying we get to \eq (10).

Finally, the term $T_5$ (\eq (9)) is given by:
\begin{align*}
T_5 & = \expectation{Q}{\sum_{m=1}^M \log q(\y_m)}
\\& = \expectation{q(\u, \lambdamax, M)}{\sum_{m=1}^M \expectation{q(\y_m)}{\log q(\y_m)}}
\\& = \expectation{q(\u, \lambdamax, M)}{M} \expectation{q(\y_m)}{\log q(\y_m)}
\\& = \expectation{q(\u) q(\lambdamax)}{\lambdamax \parameterM}\expectation{q(\y_m)}{\log q(\y_m)}
\\& = \frac{\alpha}{\beta}\expectation{q(\y_m)}{\log q(\y_m)} \expectation{q(\u)}{\parameterM}
\end{align*}

\section{Performance metrics}

We test the algorithms evaluating the $\lnorm$ norm to the true intensity function (in the synthetic settings), the test log likelihood ($\ltest$) on the test set and the negative log predicted likelihood (\nlpl) on the training set. These metrics are computed as follow:

\begin{align}
	\lnorm = \int_{\region}(\lambda(\x) - \bar{\lambda}(\x))^2d\x
\end{align}
where $\lambda(\x)$ is the true intensity function, $ \bar{\lambda}(\x)$ is the posterior mean intensity and the integral is evaluated numerically. 

\begin{align}
	\ltest = \expectation{q(\lambdamax)q(\f)}{\log \left[\exp\left(-\int_{\region}\lambda(\x)d\x\right)\prod_{\x \in \mathcal{D_{\text{test}}}}\lambda(\x)\right]}
\end{align}

where again the integral is computed via numerical integration. 

The \nlpl is computed as 
\begin{align}
	\nlpl =  - \frac{1}{S}\sum_{s=1}^{S}\log p(N_{\text{train}}|\int_{\region}\lambda^s(\x)d(\x))
\end{align}

where $S$ denotes the number of samples from the variational distributions $q(\f)$ and $q(\lambdamax)$. 

Finally, the \coverage is computed by evaluating the coverage of the \credible{s} of the posterior ($p(N | \dataset)$) and predictive ($p(N^{*} | \dataset)$). To construct the empirical count distribution we sample from the variational distributions $q(\f)$ and $q(\W)$, obtain samples of $\lambda(\x)$ and simulate $N$ or $N^*$ from $\text{Poisson}(\lambdamax  \int_{\region}\sigma(\f(\x))d\x)$.

\section{Additional experimental results}
For all comparisons we consider a \gptext with squared-exponential covariance function with equally set hyperparameters. Denote by $\hyperparam_i = (l, \sigma^2_f)$ the values of the hyperameters for the kernel function $K(\x, \x')$ on $\lambda_i(\x)$ where $l$ indicates the lenghtscale. We set:
\begin{itemize}
	\item $\hyperparam_1 = (10,1)$
	\item $\hyperparam_2 = (0.25,1)$
	\item $\hyperparam_3 = (15,1)$
\end{itemize}

For the real-world settings we have:
\begin{itemize}
	\item $\hyperparam_{\text{neuronal data}}= (10,1)$
	\item $\hyperparam_{\text{taxi data}}= (0.3,1)$
\end{itemize}

\subsection{Synthetic data experiments}
In \tbl \ref{tab:coverage_appendix_1}, \ref{tab:coverage_appendix_2} and \ref{tab:coverage_appendix_3} we report the values of \coverage for different \credible{s} and on both the training and test set.

\begin{table}
	\caption{$\lambda_1(\x)$ - \coverage performance on training and test dataset. Higher values are better.} 
	\begin{center}
\resizebox{\columnwidth}{!}{\begin{tabular}{lcccccccccccccccccc}
	\toprule
	&  \multicolumn{5}{c}{$\lambda_1(\x)$ - In-sample \coverage}    &  \multicolumn{5}{c}{$\lambda_1(\x)$ - Out-of-sample \coverage}   \\
	\cmidrule(lr){2-6}
	\cmidrule(lr){7-11}
	
	& 10\% \credible             & 20\%    \credible             & 30\% \credible               & 40\%  \credible              & 50\%        \credible        & 10\%      \credible              & 20\%        \credible            & 30\%       \credible             & 40\%     \credible              & 50\%     \credible               \\
	\cmidrule(lr){2-6}
	\cmidrule(lr){7-11}
	\our   & \textbf{1.00}              &  \textbf{1.00}              &  \textbf{1.00}              &  \textbf{1.00}             &  \textbf{1.00}             & 0.96                 & 0.88                 & \textbf{0.81}                 & \textbf{0.72}                 & \textbf{0.60}                  \\
	& (0.00)            & (0.00)             & (0.00)            & (0.00)            & (0.00)            & (0.24)               & (0.24)               & (0.23)               & (0.29)               & (0.29)               \\
	\donner &  \textbf{1.00}              &  \textbf{1.00}               &  \textbf{1.00}              &  \textbf{1.00}              &  \textbf{1.00}              & 0.95                 & 0.80                  & 0.76                 & 0.61                 & 0.52                 \\
	& (0.00)            & (0.00)             & (0.00)            & (0.00)            & (0.00)            & (0.00)                & (0.00)                & (0.00)                & (0.00)                & (0.00)                \\
	\lloyds &  \textbf{1.00}              &  \textbf{1.00}               &  \textbf{1.00}              &  \textbf{1.00}              & 0.10              & \textbf{1.00}                  & \textbf{0.97}                 & 0.75                 & 0.41                 & 0.04                 \\
	& (0.00)            & (0.00)             & (0.00)            & (0.00)            & (0.30)            & (0.00)                & (0.05)               & (0.21)               & (0.25)               & (0.09)               \\
	\adams  &  \textbf{1.00}              &  \textbf{1.00}               &  \textbf{1.00}              &  \textbf{1.00}              & 0.60              & 0.75                 & 0.60                  & 0.39                 & 0.27                 & 0.08                 \\
	& (0.00)            & (0.00)             & (0.00)            & (0.00)            & (0.49)           & (0.29)               & (0.33)               & (0.28)               & (0.22)               & (0.12)               \\
	\lgcp   & 0.70              & 0.00               & 0.00              & 0.00              & 0.00              & 0.48                 & 0.22                 & 0.08                 & 0.03                 & 0.01                 \\
	& (0.46)           & (0.00)             & (0.00)            & (0.00)            & (0.00)            & (0.00)                & (0.00)                & (0.00)                & (0.00)                & (0.00)                \\
	\bottomrule
\end{tabular}}
		\label{tab:coverage_appendix_1}
\end{center}
\end{table}

\begin{table}
	\caption{$\lambda_2(\x)$ - \coverage performance on training and test dataset. Higher values are better.} 
	\begin{center}
		\resizebox{\columnwidth}{!}{\begin{tabular}{lcccccccccccccccccccccc}
	\toprule
	&  \multicolumn{5}{c}{$\lambda_2(\x)$ - In-sample \coverage}    &  \multicolumn{5}{c}{$\lambda_2(\x)$ - Out-of-sample \coverage}   \\
\cmidrule(lr){2-6}
\cmidrule(lr){7-11}

	& 10\% \credible             & 20\%    \credible             & 30\% \credible               & 40\%  \credible              & 50\%        \credible        & 10\%      \credible              & 20\%        \credible            & 30\%       \credible             & 40\%     \credible              & 50\%     \credible               \\
	\hline
	\our    & \textbf{1.00}              &  \textbf{1.00}              &  \textbf{1.00}              &  \textbf{1.00}             &  \textbf{1.00}             & \textbf{1.00}                 & \textbf{0.97}                 & \textbf{0.91}                 & \textbf{0.88}                 & \textbf{0.86}                 \\
	& (0.00)            & (0.00)             & (0.00)            & (0.00)            & (0.00)            & (0.00)                & (0.09)               & (0.24)               & (0.23)               & (0.22)               \\
	\donner & \textbf{1.00}              &  \textbf{1.00}              &  \textbf{1.00}              &  \textbf{1.00}             &  \textbf{1.00}              & 0.92                 & 0.92                 & 0.89                 & 0.84                 & 0.82                 \\
	& (0.00)            & (0.00)             & (0.00)            & (0.00)            & (0.00)            & (0.00)                & (0.00)                & (0.00)                & (0.00)                & (0.00)                \\
	\lloyds & \textbf{1.00}              &  \textbf{1.00}              &  \textbf{1.00}              &  \textbf{1.00}                    & 0.10              & 0.92                 & 0.86                 & 0.76                 & 0.45                 & 0.05                 \\
	& (0.00)            & (0.00)             & (0.00)            & (0.00)            & (0.30)            & (0.24)               & (0.23)               & (0.26)               & (0.26)               & (0.05)               \\
	\adams  &\textbf{1.00}            & 0.90               & 0.70              & 0.40              & 0.30              & 0.90                  & 0.90                  & 0.64                 & 0.14                 & 0.00                  \\
	& (0.00)            & (0.30)             & (0.46)           & (0.49)           & (0.46)           & (0.00)                & (0.00)                & (0.09)               & (0.05)               & (0.00)                \\
	\lgcp   & 0.10              & 0.00               & 0.00              & 0.00              & 0.00              & 0.80                  & 0.22                 & 0.04                 & 0.00                  & 0.00                  \\
	& (0.30)            & (0.00)             & (0.00)            & (0.00)            & (0.00)            & (0.24)               & (0.16)               & (0.08)               & (0.00)                & (0.00)                \\
	\hline
\end{tabular}}
\label{tab:coverage_appendix_2}
\end{center}
\end{table}

\begin{table}
	\caption{$\lambda_3(\x)$ - \coverage performance on training and test dataset. Higher values are better.} 
	\begin{center}
		\resizebox{\columnwidth}{!}{\begin{tabular}{lcccccccccccccccccccccc}
	\toprule
	&  \multicolumn{5}{c}{$\lambda_3(\x)$ - In-sample \coverage}    &  \multicolumn{5}{c}{$\lambda_3(\x)$ - Out-of-sample \coverage}   \\
\cmidrule(lr){2-6}
\cmidrule(lr){7-11}

& 10\% \credible             & 20\%    \credible             & 30\% \credible               & 40\%  \credible              & 50\%        \credible        & 10\%      \credible              & 20\%        \credible            & 30\%       \credible             & 40\%     \credible              & 50\%     \credible               \\
	\hline
	\our    &  \textbf{1.00}              &  \textbf{1.00}              &  \textbf{1.00}              &  \textbf{1.00}             &  \textbf{1.00}                & \textbf{1.00}                  & \textbf{1.00}                 & 0.99                 & 0.97                 & 0.92                 \\
	& (0.00)            & (0.00)             & (0.00)            & (0.00)            & (0.00)            & (0.00)                & (0.00)                & (0.00)                & (0.00)                & (0.12)               \\
	\donner &  \textbf{1.00}              &  \textbf{1.00}              &  \textbf{1.00}              &  \textbf{1.00}             &  \textbf{1.00}             & \textbf{1.00}                  & \textbf{1.00}                  & 0.97                 & 0.91                 & 0.78                 \\
	& (0.00)            & (0.00)             & (0.00)            & (0.00)            & (0.00)            & (0.00)                & (0.00)                & (0.00)                & (0.00)                & (0.00)                \\
	\lloyds &  \textbf{1.00}              &  \textbf{1.00}              &  \textbf{1.00}              &  \textbf{1.00}                    & 0.10              & 0.97                 & 0.94                 & 0.83                 & 0.43                 & 0.03                 \\
	& (0.00)            & (0.00)             & (0.00)            & (0.00)            & (0.30)            & (0.09)               & (0.15)               & (0.19)               & (0.14)               & (0.05)               \\
	\adams  & 0.80              & 0.70               & 0.50              & 0.40              & 0.00              & 0.82                 & 0.54                 & 0.49                 & 0.34                 & 0.02                 \\
	& (0.40)            & (0.46)            & (0.50)            & (0.49)           & (0.00)            & (0.12)               & (0.05)               & (0.03)               & (0.07)               & (0.04)               \\
	\lgcp   & \textbf{1.00}              &  \textbf{1.00}              &  \textbf{1.00}              &  \textbf{1.00}             &  \textbf{1.00}           & \textbf{1.00}                  & \textbf{1.00}                  & \textbf{1.00}                  & \textbf{1.00}                  & \textbf{0.95}                 \\
	& (0.00)            & (0.00)             & (0.00)            & (0.00)            & (0.00)            & (0.00)                & (0.00)                & (0.00)                & (0.00)                & (0.00)                \\
	\hline
\end{tabular}}
\label{tab:coverage_appendix_3}
\end{center}
\end{table}

\subsection{Real data experiments}

In \tbl \ref{tab:coverage_real_appendix} we report the values of \coverage for different \credible{s} and on both the training and test set.

\begin{table}
	\caption{Real data. Values are given as In-sample - Out-of-sample \coverage. Mean and standard errors (in parenthesis) are computed across different seeds.} 
	\begin{center}
		\begin{tabular}{@{\extracolsep{4pt}}@{\kern\tabcolsep}lcccccccccccccccc}
			%{llllllrrrrrrrrrr}
			\toprule
			& \multicolumn{5}{c}{Neuronal data} \\
			\cmidrule(lr){2-6}
			&  10\%  \coverage&  20\% \coverage& 30\%  \coverage&  40\% \coverage& 50\%\coverage \\
			\hline
			\multirow{2}{*}{\our}    &  \textbf{1.00}-\textbf{1.00} &  \textbf{1.00}-\textbf{1.00} &  \textbf{1.00}-\textbf{1.00} & \textbf{0.99}-\textbf{0.56}      &  \textbf{0.01}- 0.00  \\
			&  0.00 - 0.00 & 0.00 - 0.00 &  0.00-0.00& (0.10)-(0.50) &  (0.10)-0.00 \\
			\multirow{2}{*}{\donner}   &   \textbf{1.00}-\textbf{1.00} &  \textbf{1.00}-0.62 &  \textbf{1.00}-0.03 & 0.78-0.00      &  0.00 - 0.00  \\
			& 0.00 - 0.00 &  0.00-(0.49)&  0.00-(0.17) & (0.41)-0.00 &  0.00 - 0.00 \\
			\multirow{2}{*}{\lloyds}   &    \textbf{1.00}-0.53 &  \textbf{1.00}-0.00 &  \textbf{1.00}-0.00 & 0.83-0.00    & \textbf{0.01}-0.00    \\
			& 0.00-(0.50) &  0.00 - 0.00&  0.00 - 0.00& (0.38)-0.00  &  (0.10)-0.00 \\
			\midrule
			& \multicolumn{5}{c}{Taxi data}\\
			\cmidrule(lr){2-6}
			&  10\% \coverage &  20\%\coverage & 30\%  \coverage&  40\% \coverage& 50\% \coverage\\
			\hline
			\multirow{2}{*}{\our}    &  \textbf{1.00}-1.00      &  \textbf{1.00}-\textbf{1.00} & 0.81-\textbf{0.37}     & 0.09-\textbf{0.01}     &          0.00-0.00 \\
			&   0.00-0.00   &  0.00-0.00 & (0.39)-(0.48) & (0.29)-(0.10) &       0.00-0.00 \\
			\multirow{2}{*}{\donner}   &   0.49-0.93   &  0.00-0.13  & 0.00-0.00       & 0.00-0.00        &          0.00-0.00  \\
			&  (0.50)-(0.26) &  0.00-(0.34) & 0.00-0.00      &0.00-0.00       &    0.00-0.00 \\
			\multirow{2}{*}{\lloyds}     & \textbf{1.00}-0.00      &  \textbf{1.00}-0.00 & \textbf{0.98}-0.00     & \textbf{0.48}-0.00     &          0.00-  0.00  \\
			&  0.00- (0.00)     &  0.00-(0.00) & (0.14)-(0.00)     & (0.50)-0.00   &          0.00-0.00 \\
			\bottomrule
		\end{tabular}
	\label{tab:coverage_real_appendix}
	\end{center}
\end{table}

%\bibliographystyle{apalike}
%\bibliography{references}